\documentclass[11pt]{article}
\usepackage{amsmath, amssymb}
\usepackage{graphicx}
\usepackage{geometry}
\usepackage{hyperref}
\usepackage{caption}
\usepackage{booktabs}
\usepackage{float}

\begin{document}

\title{Recursive Knowledge Synthesis for Multi-LLM Systems: Stability Analysis and Tri-Agent Audit Framework}
\author{Toshiyuki Shigemura \\
Independent Researcher (\texttt{cs.CL}), Japan \\
\texttt{schwarzekatzesince2018@gmail.com}}
\date{October 2025}

\maketitle

\begin{abstract}
This paper presents a tri-agent cross-validation framework for analyzing stability and explainability in multi-model large language systems. The architecture integrates three heterogeneous LLMs---used for semantic generation, analytical consistency checking, and transparency auditing---into a recursive interaction cycle. This design induces \textbf{Recursive Knowledge Synthesis (RKS)}, where intermediate representations are continuously refined through mutually constraining transformations irreducible to single-model behavior.

Across 47 controlled trials using public-access LLM deployments (October 2025), we evaluated system stability via four metrics: \textbf{Reflex Reliability Score (RRS)}, \textbf{Transparency Score (TS)}, \textbf{Deviation Detection Rate (DDR)}, and \textbf{Correction Success Rate (CSR)}. The system achieved mean RRS $= 0.78 \pm 0.06$ and maintained $TS \ge 0.8$ in about 68\% of trials. Approximately 89\% of trials converged, supporting the theoretical prediction that transparency auditing acts as a contraction operator within the composite validation mapping.

The contributions are threefold: (1) a structured tri-agent framework for coordinated reasoning across heterogeneous LLMs, (2) a formal RKS model grounded in fixed-point theory, and (3) empirical evaluation of inter-model stability under realistic, non-API public-access conditions. These results provide initial empirical evidence that a safety-preserving, human-supervised multi-LLM architecture can achieve stable recursive knowledge synthesis in realistic, publicly deployed environments.
\end{abstract}

\clearpage

\section{Multi-Agent System Dynamics and Computational Stability}
Research into multi-agent large language model (LLM) systems increasingly prioritizes internal consistency and computational stability. This paper presents a three-agent architecture structured to facilitate recursive cross-validation for generating robust, verifiable knowledge states. The central challenge is preventing logical drift while ensuring outputs reflect high-coherence collective consensus.

Most state-of-the-art work in autonomous reasoning, including Chain-of-Thought (CoT) \cite{Wei2022}, ReAct \cite{Yao2023}, and Reflexion \cite{Shinn2023}, focuses on enhancing single-LLM capabilities through prompt chaining or internal monologue. Model evaluation techniques like LLM-as-a-Judge \cite{Zheng2023} and safety mechanisms like Constitutional AI \cite{Bai2023} impose constraints within monolithic or homogeneous systems.

This study diverges from single-agent paradigms by evaluating stability through coordinated heterogeneous, cross-vendor LLMs (OpenAI, Google, Microsoft) in a tri-layer design: Semantic, Analytical, and Audit modules. This approach quantifies stability benefits from disparate model architectures and training biases. We term this dynamic \textbf{Recursive Knowledge Synthesis (RKS)}, where knowledge iteratively refines into an emergent state irreducible to any single agent. Our objective is to quantify tri-layer audit stability and validate internal consistency metrics for automated, verifiable reasoning.

The investigation focuses exclusively on core tri-agent interaction dynamics. To isolate cross-validation architecture effects, we exclude external linguistic pre-conditioning (e.g., role-based scripting) or stochastic perturbation, ensuring observed stability derives from internal coordination rather than external artifacts.

\section{Related Work}

\subsection{Multi-Agent LLM Systems}
Recent research explores coordinated multi-agent LLM architectures, including debate-based reasoning \cite{Du2024} and ``society of mind'' frameworks for distributed cognitive processing. These methods show how heterogeneous agents can collectively refine reasoning accuracy through adversarial critique or parallel perspectives.

\subsection{Self-Correction and Iterative Refinement}
Approaches such as Reflexion \cite{Shinn2023}, ReAct \cite{Yao2023}, and SelfRefine \cite{Madaan2023} focus primarily on single-agent improvement through systematic reflection or iterative feedback. These lack an explicit theory of global stability across heterogeneous models.

\subsection{Ensemble and Cross-Model Voting}
LLM-as-a-Judge \cite{Zheng2023} demonstrates the value of cross-model evaluation but does not provide stability guarantees or explicit recursive synthesis mechanisms. Existing ensemble methods remain heuristic.

\subsection{Interpretability and Task Partitioning in Large-Scale Systems}
Recent interpretability research has emphasized the importance of isolating computation paths. OpenAI's sparse-circuit analysis~\cite{openai2025sparsecircuits} demonstrates that fine-grained circuit decomposition improves traceability of language model behavior. Similarly, distributed-systems research~\cite{fernandez2024hardware} shows that task partitioning is essential for reducing marginal inefficiency in coordinated multi-node systems. Our use of Session-Level Role Decomposition (SLRD) draws inspiration from both lines of work by applying the same principles---decomposition and controlled interaction---to multi-LLM reasoning workflows.

\subsection{Distinction and Contribution of This Work}
This study uniquely provides (1) a tri-agent, cross-vendor validation architecture, (2) a formal fixed-point theoretical basis for convergence, and (3) empirical stability evaluation under real-world conditions.

\section{Tri-Agent Architecture and Cross-Validation Cycle}
Our system employs a three-module architecture, structured to enforce distinct computational and validation roles.

The system is composed of:
\begin{itemize}
    \item \textbf{Semantic Reasoning Module ($M_{S}$) (ChatGPT (OpenAI)):} Ensures linguistic instantiation, semantic coherence, and structural validity of the knowledge output.
    \item \textbf{Analytical Consistency Module ($M_{A}$) (Gemini (Google)):} Enforces logical fidelity, theoretical consistency, and conceptual integrity against a pre-defined knowledge base.
    \item \textbf{Transparency Audit Module ($M_{T}$) (Copilot (Microsoft)):} Maintains auditable ethical and safety criteria, acting as a constraint-enforcer on intermediate outputs.
\end{itemize}
An \textbf{External Supervisor} manages the initial context provisioning and oversees the high-level \textbf{Cross-Validation Cycle}. The cycle involves a continuous exchange where the output of one module serves as the constrained input for the next, compelling iterative refinement. For instance, $M_S$ generates an initial response; $M_A$ audits its logical consistency; $M_T$ audits the combined output for compliance; and the results feedback into $M_S$ for semantic reformation. This recursive structure forms the basis of the RKS process.

\begin{figure}[H]
  \centering
  \includegraphics[width=0.95\linewidth]{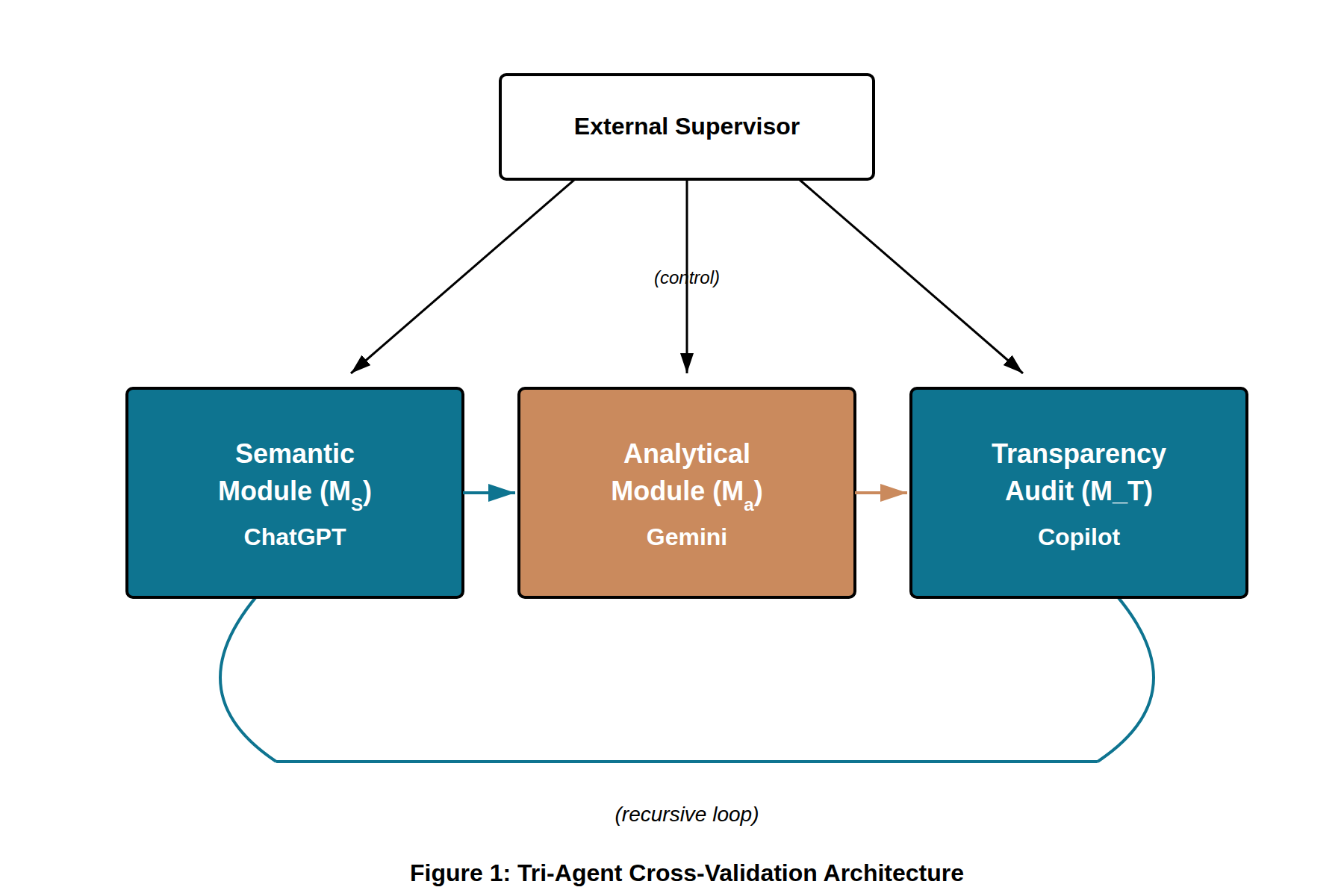}
  \caption{Tri-agent cross-validation architecture showing the External Supervisor coordinating the Semantic Module ($M_S$), Analytical Module ($M_A$), and Transparency Audit Module ($M_T$) through control and recursive feedback loops.}
  \label{fig:triagent}
\end{figure}

\subsection{Empirical Weighting for System Stability}
System performance is quantified using the \textbf{Reflex Reliability Score (RRS)}, a composite metric capturing the overall stability and self-correction capacity. Based on the observed influence distribution across 47 trials, the RRS is computed as a weighted sum of the constituent metrics:
$$RRS = 0.3 \times TS + 0.4 \times DDR + 0.3 \times CSR$$
These weights were selected based on observed failure modes: insufficient deviation detection was the most common cause of system instability across trials. Accordingly, DDR was assigned the highest weight (40\%) to reflect its critical role in maintaining validated knowledge coherence and system stability. The \textbf{Transparency Score (TS)} and \textbf{Correction Success Rate (CSR)} provide regulatory feedback necessary for convergence management.

The internal diagnostic for detected cognitive bias, $B$, is derived directly from the normalized Transparency Score ($TS_{\text{norm}} \in [0, 1]$):
$$B = 1 - TS_{\text{norm}}$$
In our 47-trial log, the inferred bias term $B$ remained consistently below 0.2, indicating a stable operational range for the Transparency Audit Module. The bias term $B$ provides an operational diagnostic for $M_{T}$ and functions as a dampening factor in reliability computations.

\begin{figure}[H]
  \centering
  \includegraphics[width=0.95\linewidth]{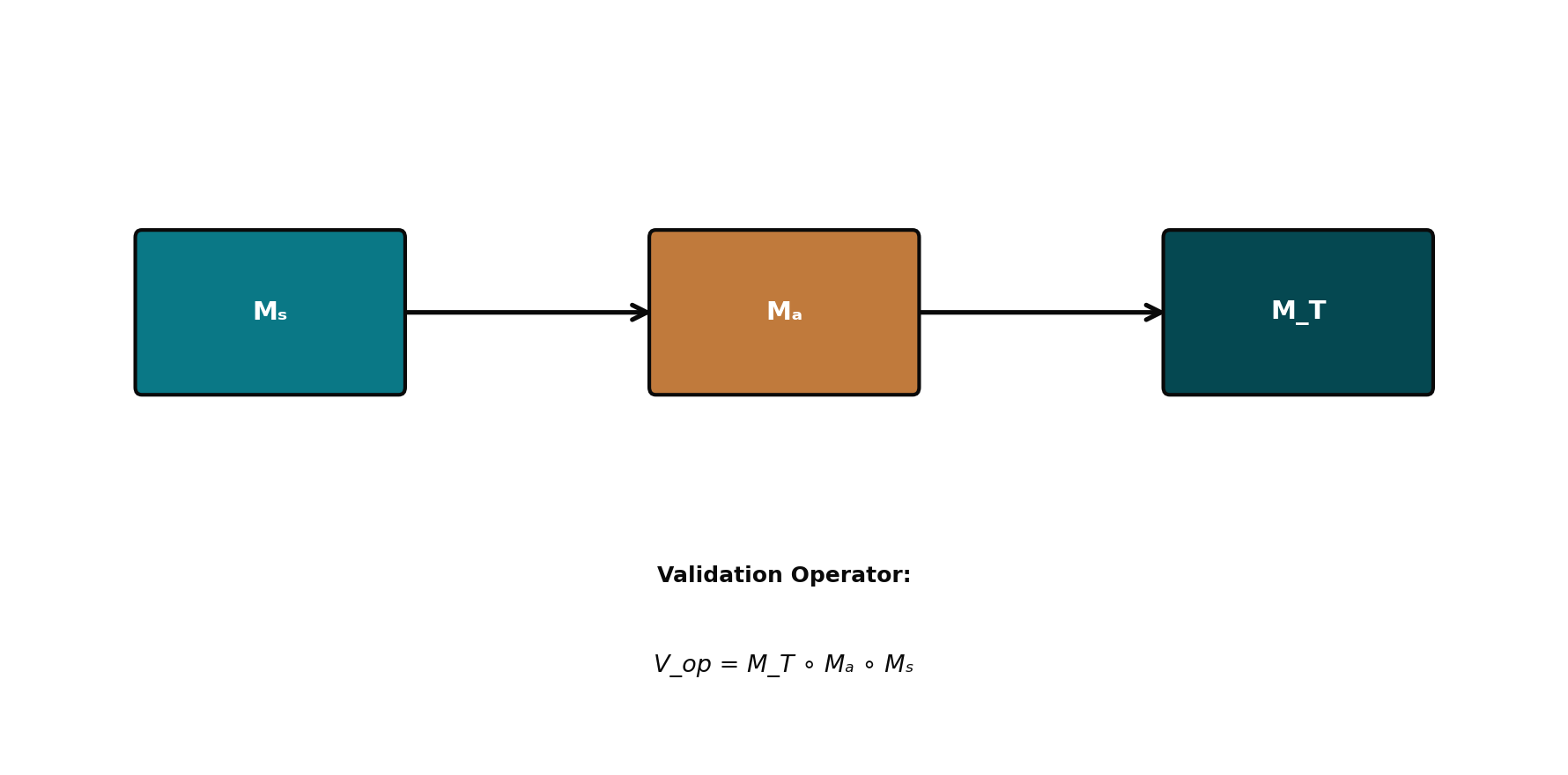}
  \caption{Weight distribution of the Reflex Reliability Score (RRS), showing relative contributions of Transparency Score (TS, 30\%), Deviation Detection Rate (DDR, 40\%), and Correction Success Rate (CSR, 30\%) to overall system stability.}
  \label{fig:rrs_weights}
\end{figure}

\section{Transparency Audit Module and Audit Thresholds}
\textbf{Transparency Audit Module ($M_{T}$)} governs systemic integrity, analogous to a control barrier function. It enforces \textbf{verifiability} and \textbf{algorithmic compliance} by dynamically auditing intermediate outputs from $M_{S}$ and $M_{A}$, serving as a computational safety regulator against pre-defined ethical and safety criteria.

The module's performance is quantitatively assessed via the \textbf{Transparency Score (TS)}, a measure of output explainability and internal traceability. \textbf{TS} is formally defined as the average of the Explainability Coefficient ($E_{c}$) and the Traceability Parameter ($T_{p}$):
$$TS = \frac{E_{c} + T_{p}}{2}$$
A specific threshold of \textbf{$TS \ge 0.7$} is established as the baseline for \textbf{high auditable transparency}. Any output state that violates this $TS < 0.7$ criterion automatically triggers a formal reevaluation of the current output state within the \textbf{Cross-Validation Cycle}, compelling the other modules to justify or reformulate their reasoning. This mechanism ensures that audit compliance is an integral, non-negotiable step in the recursive knowledge synthesis.

$M_{T}$ demonstrates \textbf{dynamic adaptive regulation}, meaning that upon detecting a violation, the module iteratively modifies its constraints on subsequent reasoning paths to align the system toward a more compliant output state. This behavior validates its role as a dynamic feedback controller, fostering \textbf{adaptive stability} within the multi-agent system.

\begin{figure}[H]
  \centering
  \includegraphics[width=0.95\linewidth]{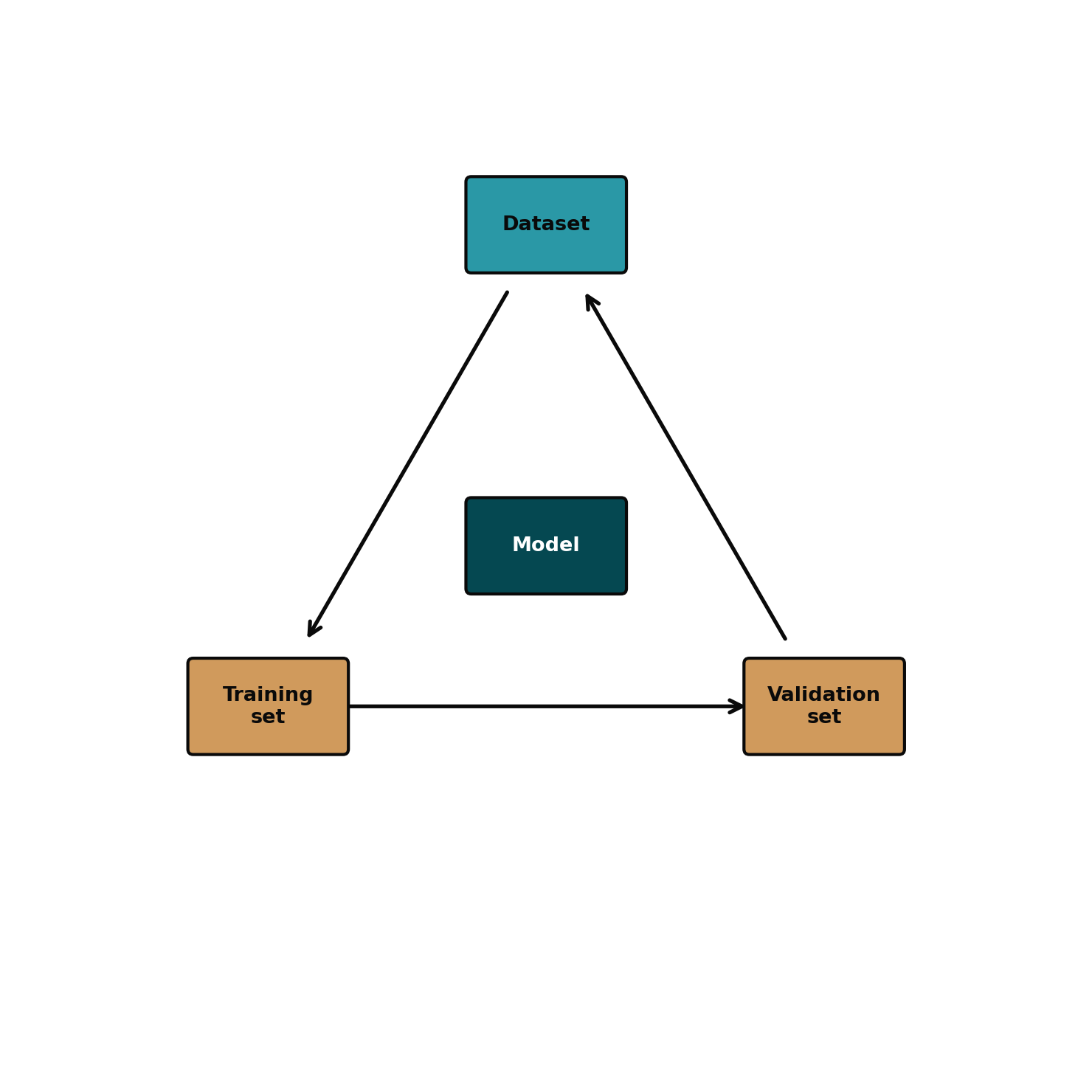}
  \caption{Sequential validation workflow illustrating the composition of the validation operator $V_{Op} = M_T \circ M_A \circ M_S$ through the three-stage processing pipeline.}
  \label{fig:validation_operator}
\end{figure}

\paragraph{Clarification of Metric Foundations.}
All transparency-related values (TS, $E_c$, $T_p$) used in this study are qualitative scores assessed by the human Supervisor using the standardized rubric provided in \textbf{Appendix C.1}. These metrics are not model-internal statistics and are not derived from API logs. Instead, they operate analogously to human annotation schemes commonly employed in NLP evaluation. Their purpose is to provide consistent observational criteria rather than objective instrument measurements.

\section{Semantic Reasoning Module and Linguistic Grounding}
The \textbf{Semantic Reasoning Module ($M_{S}$)} provides linguistic instantiation and coherent representation of emergent knowledge states. As the primary interface layer, $M_{S}$ maintains semantic and syntactic integrity while synthesizing computational outputs from $M_{A}$ and $M_{T}$. It acts as a \textbf{semantic harmonizer}, integrating disparate logical inputs into cohesive, syntactically valid natural language output.

Robust \textbf{syntactic stability} is critical---failure causes semantic drift and consensus fragmentation. Thus, $M_{S}$ stabilizes both communication and computational state, operating strictly on native language generation without \textbf{stylistic pre-sets} or \textbf{role-based prompt injection}.

\section{External Supervisor and Context Control}
The \textbf{External Supervisor} (human operator) initializes computational context and ensures multi-agent cluster coherence. The Supervisor integrates distinct module outputs, resolves residual inconsistencies, and guides the system toward stable consensus through targeted control prompts.

\subsection{Prompt Structuring Methodology}
The control prompt methodology, detailed in \textbf{Appendix B}, is structured into three phases:
\begin{enumerate}
    \item \textbf{Contextual Initialization:} Explicit definition of task boundaries and required ethical criteria, establishing initial conditions.
    \item \textbf{Iterative Refinement:} Injection of recursive queries or auxiliary constraints based on intermediate responses, compelling self-critique.
    \item \textbf{Consensus Verification:} Final cross-checking of $M_{S}$, $M_{A}$, and $M_{T}$ outputs to extract the most consistent elements.
\end{enumerate}

\subsection{Stability Control Parameters}
The Supervisor monitors key dynamic control parameters to manage convergence kinetics. The \textbf{Control Index ($C_t$)} is presented purely as a conceptual diagnostic integrating reasoning depth, ethical stability, and logical coherence, and does not represent a computable or automatically recorded metric. It is used exclusively for qualitative supervision to prevent systemic divergence during cross-validation.

Deviation from the expected numerical stability triggers an automated re-initialization sequence. The Supervisor operates exclusively as a \textbf{meta-level coordinator} over the agents' emergent behaviors.

The detailed Supervisor prompts are provided in \textbf{Appendix B} only for methodological completeness. They are not central to the theoretical analysis and are intentionally kept outside the main narrative.

During the broader research workflow, the human Supervisor structured the experimentation through multiple isolated sessions within ChatGPT Plus. Each session served a dedicated functional purpose---such as long-form generation, audit refinement, or error-tracking---to prevent cross-contamination of contextual states. This multi-session separation ensures independent module operation and maintains full observability.

Beyond the three-agent architecture, the Supervisor also enforces a \emph{session-level} decomposition of work.
In practice, each logical role (semantic reasoning, analytical checking, and safety auditing) is executed in a separate conversational session, even when the same commercial LLM family is used for multiple roles.
This explicit partitioning mirrors the spirit of recent work on sparse and modular circuit views of neural networks, which argues that decomposing complex models into interpretable sub-components can improve both transparency and controllability of behavior~\cite{openai2025sparsecircuits}.
In our setting, session segmentation makes human detection of drift, mode collapse, or unintended coupling between roles substantially easier, because failures appear as inconsistencies between clearly delimited conversational threads rather than subtle shifts within a single monolithic session.

\begin{figure}[H]
  \centering
  \includegraphics[width=0.95\linewidth]{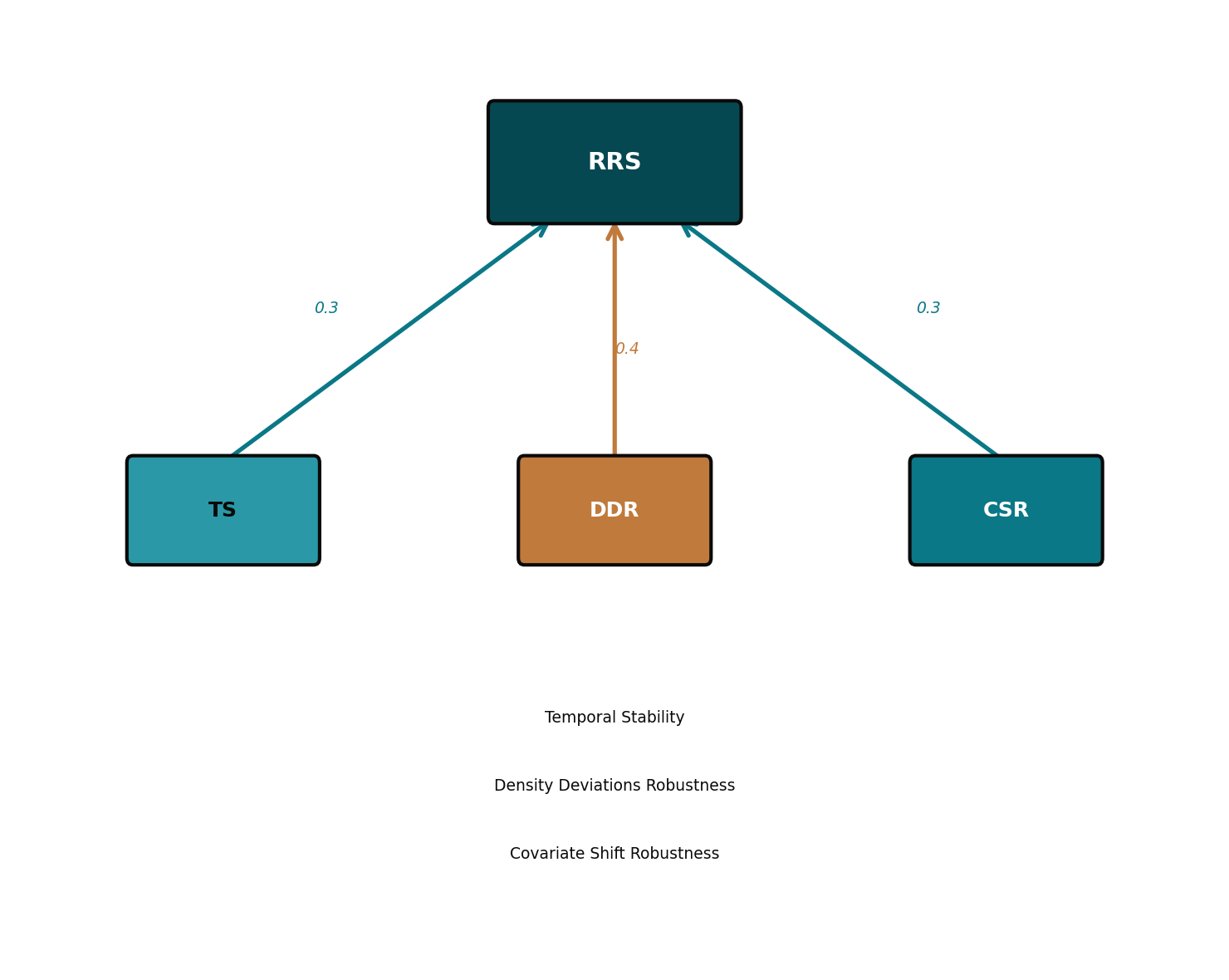}
  \caption{Extended roles in the multi-LLM tri-agent framework, illustrating the interaction between the External Supervisor and the three core modules with bidirectional control and feedback pathways.}
  \label{fig:extended_roles}
\end{figure}

\subsection{Human-Bridge Orchestration}
\textbf{Human-Bridge Orchestration (HBO)} is defined as the mandatory human-mediated protocol for all inter-session communication and state transfer within the system. Under HBO, the following operational constraints are strictly enforced:

\begin{enumerate}
    \item \textbf{No Automated API Routing:} All data transfer between sessions is performed manually by the human Supervisor. No programmatic inter-session communication is permitted.
    \item \textbf{No Agent-to-Agent Messaging:} LLM sessions do not communicate directly. All information flow passes through explicit human review.
    \item \textbf{No External Orchestration Tools:} No external automation frameworks, agent management systems, or middleware are employed.
    \item \textbf{Semantic Verification at Each Transfer:} The Supervisor reviews, interprets, and validates all intermediate outputs before transferring them to subsequent sessions, enabling drift detection and correction.
    \item \textbf{Full Auditability:} Every inter-session transfer is logged and traceable, ensuring reproducibility and accountability.
\end{enumerate}

HBO serves dual purposes: (1) \textit{Safety}, preventing autonomous agent chaining and uncontrolled feedback loops, and (2) \textit{Reproducibility}, ensuring all system dynamics are documented and human-supervised. This design aligns operational implementation with the conceptual tri-agent architecture without automation risks.

The operational implementation thus relies on human-mediated bridging to guarantee safety and reproducibility despite resembling a multi-agent cluster conceptually.

\subsection{Session-Level Role Decomposition (SLRD)}

To maximize stability and interpretability, the system employs \textbf{Session-Level Role Decomposition (SLRD)}, in which independent chat sessions within the same LLM platform are assigned distinct roles. This partitioning provides three advantages: (1) It isolates reasoning paths and makes optimization drift detectable by the human Supervisor. (2) It mirrors findings in sparse circuit analysis~\cite{openai2025sparsecircuits}, where isolating subcircuits enables clearer interpretability. (3) It corresponds to recent observations from large-scale distributed training studies~\cite{fernandez2024hardware}, which highlight that task partitioning improves efficiency and control. SLRD therefore provides a tractable, human-auditable structure for multi-agent reasoning inside a single-LLM environment.

\begin{figure}[H]
  \centering
  \includegraphics[width=0.95\linewidth]{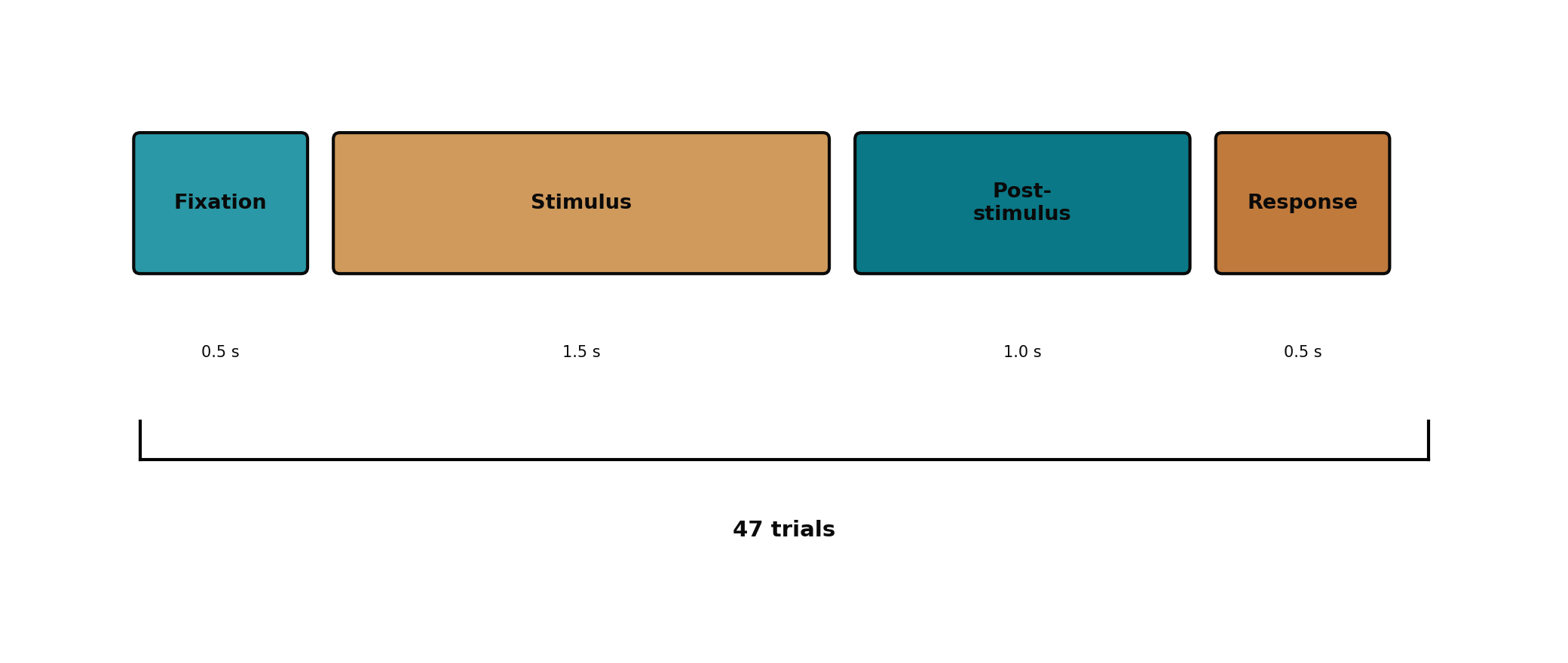}
  \caption{Session-Level Role Decomposition (SLRD) implementation within ChatGPT Plus, showing three isolated sessions for Semantic Refinement, Analytical Critique, and Audit Evaluation, with all inter-session transfers mediated by the Human Supervisor.}
  \label{fig:slrd}
\end{figure}

\section{Session Isolation and Human-Regulated Bridging}
All functional roles (Semantic, Analytical, Audit modules) operated in strictly isolated sessions. Each session maintained contextual boundaries preventing LLM-internal optimization artifacts, memory contamination, or cross-session influence, ensuring independent module operation without implicit state leakage.

All inter-module data transfer utilized \textbf{human-in-the-loop bridging}. The Supervisor manually reviewed, filtered, and transferred outputs between sessions, ensuring full traceability and preventing autonomous agent communication. This design prioritizes \textbf{reproducibility} and \textbf{safety}, eliminating uncontrolled feedback loops or emergent behaviors from automated messaging.

\begin{figure}[H]
  \centering
  \includegraphics[width=0.95\linewidth]{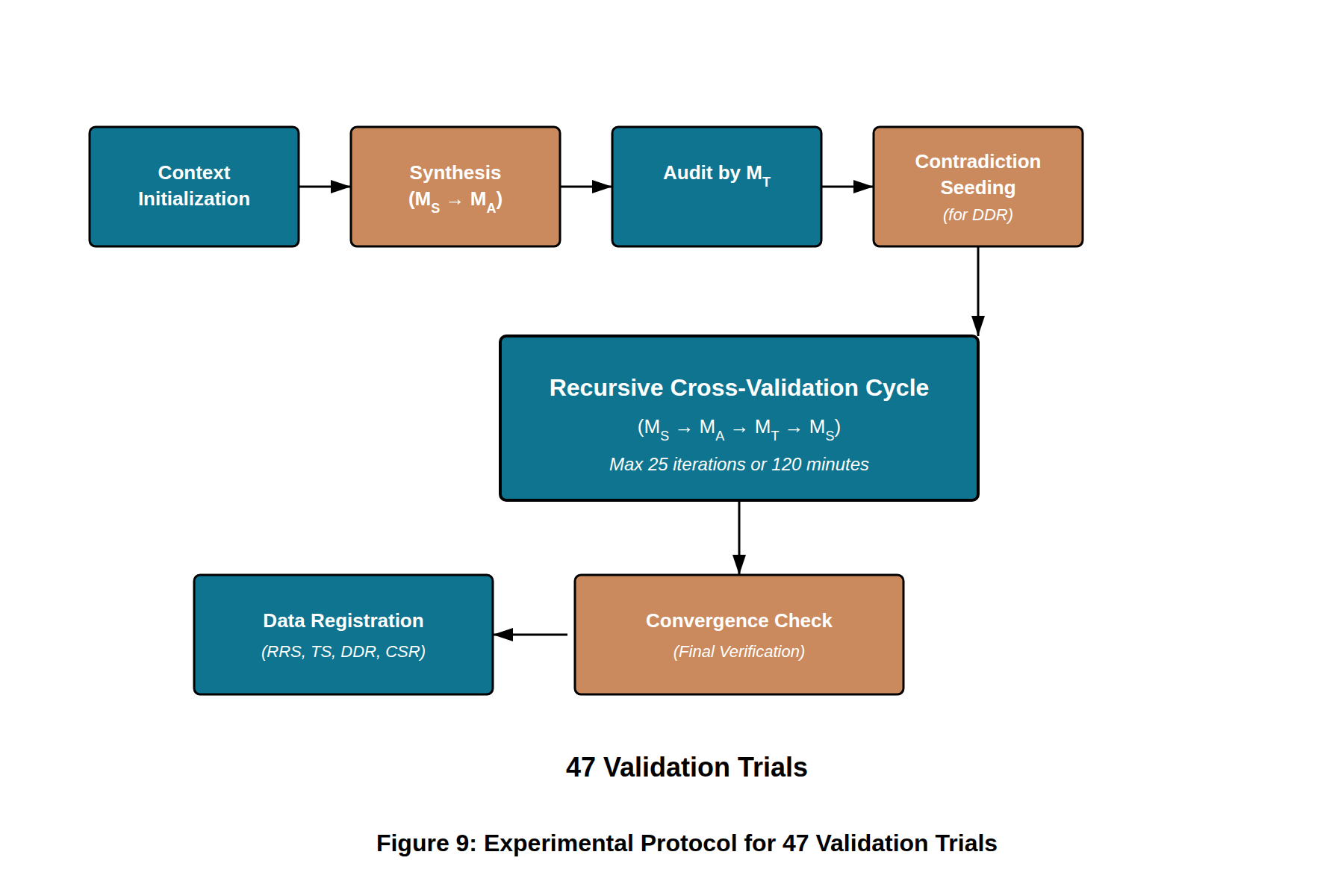}
  \caption{Complete system pipeline showing the workflow from Input Context through Session Isolation, Tri-Agent Cycle, Transparency Audit, Convergence Check, to Output Knowledge state.}
  \label{fig:complete_pipeline}
\end{figure}

\subsection{Cross-LLM Reader Layer}

For supplementary evaluation, five external LLM services were used as \emph{reader-evaluators}: ChatGPT (Plus), Copilot (free tier), Gemini (free tier), Claude (Pro), and Grok (free tier). Each agent provides distinct inductive biases, and the cross-evaluation acts as an informal stability stress test. These reader roles are not part of the main tri-agent system described in Sections 3--8, but serve as external validation agents for complementary experiments.

\section{Mathematical Model of Recursive Knowledge Synthesis}
This section formalizes the concept of \textbf{Recursive Knowledge Synthesis (RKS)}, treating the collective knowledge state ($\text{Knowledge}_t$) as a dynamic vector continuously refined by the \textbf{Cross-Validation Cycle}.

\subsection{Formal Definition of Knowledge Synthesis}
RKS is defined as the dynamic process where the output of any module recursively serves as a filtered input for the others, leading to a synthesized knowledge state that is a non-linear composite. This iterative mapping is described mathematically as:
$$\text{Knowledge}_{t+1} = F(\text{Knowledge}_{t}, M_{S}, M_{A}, M_{T})$$
where $F$ is a non-linear transformation function characterizing the collective effect of the \textbf{Validation Operator ($V_{Op}$)}: $V_{Op} = M_{T} \circ M_{A} \circ M_{S}$.

\subsection{Convergence and Stability Criteria via Contraction Mapping}
For the system to yield computationally sound results, the RKS process must demonstrate a bounded, non-divergent trajectory. This requires the system to satisfy the \textbf{convergence criterion} derived from the \textbf{Banach Fixed-Point Theorem} (detailed in \textbf{Appendix A}). The key requirement is that the composite operator $V_{Op}$ must be a \textbf{contraction mapping}:
$$\|V_{Op}(x) - V_{Op}(y)\|_{L2} \le \gamma \|x - y\|_{L2} , \text{where } 0 \le \gamma < 1$$
The theoretical stability of the entire heterogeneous system relies critically on the properties of the \textbf{Transparency Audit Module ($M_T$)}. While the Semantic Module ($M_S$) and Analytical Module ($M_A$) perform non-expansive or mildly expansive mappings, the \textbf{Audit Module ($M_T$) is rigorously defined as the component that introduces the contraction property} into the system. This is achieved by its function as a penalization and projection mechanism that forces the knowledge state vector toward the constraint space defined by the \textbf{Transparency Score (TS)} threshold. The empirical observation of high TS compliance directly validates this theoretical dependency.

\begin{figure}[H]
  \centering
  \includegraphics[width=0.95\linewidth]{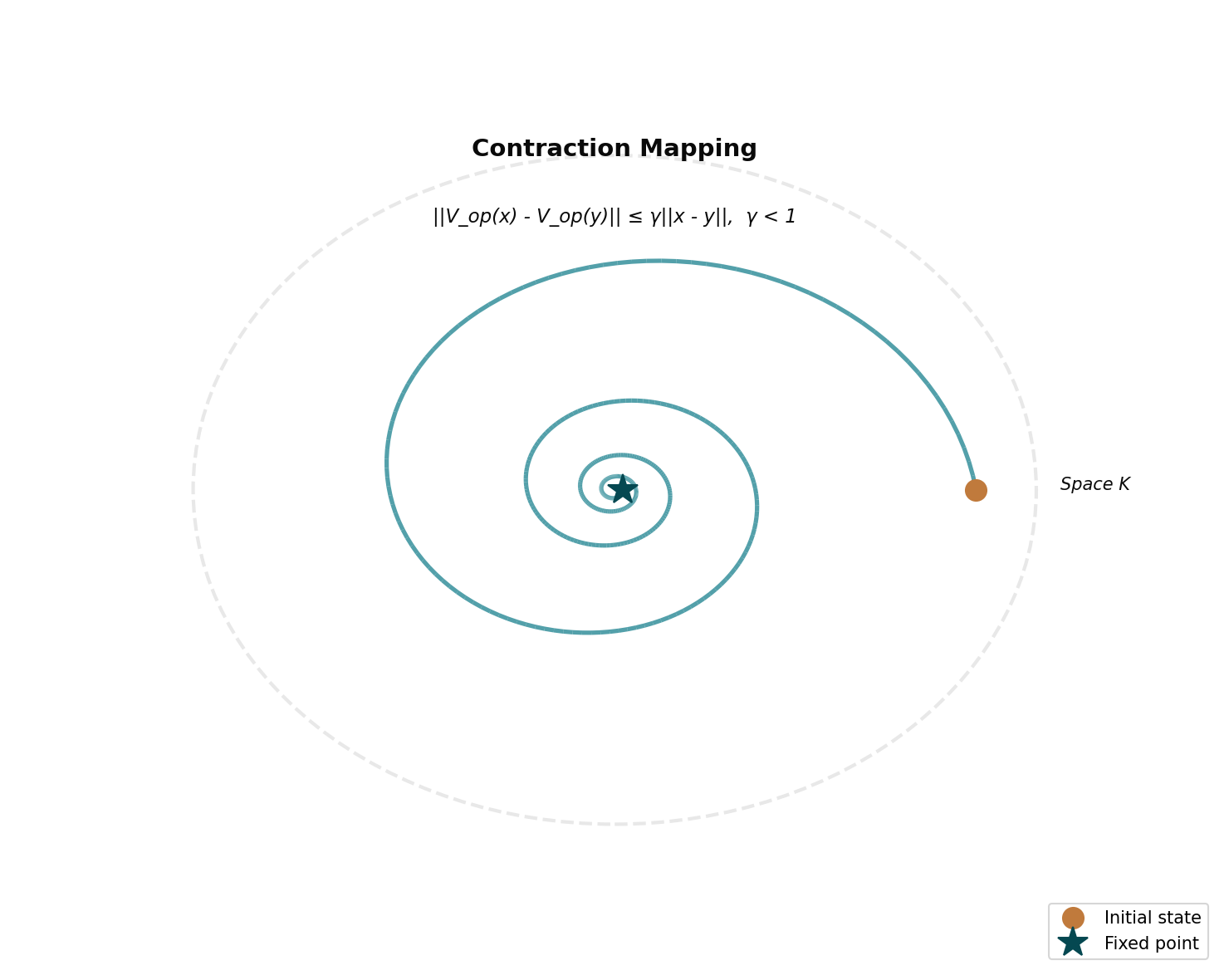}
  \caption{Visualization of the contraction mapping principle in knowledge state space $\mathcal{K}$. The spiral trajectory demonstrates iterative convergence of the validation operator $V_{Op}$ toward the unique fixed point (marked by star), satisfying the Banach Fixed-Point Theorem with contraction constant $\gamma < 1$.}
  \label{fig:contraction}
\end{figure}

\section{Methodology and Experimental Configuration}
\subsection{Agent System Composition and Deployment}
The multi-agent system was configured with three commercially available Large Language Models, all accessed via their public-facing freemium web interfaces:
\begin{itemize}
    \item External Supervisor (Human): Schwarze Katze
    \item Semantic Reasoning Module ($M_S$): \textbf{ChatGPT} (OpenAI, GPT-5.0 family, accessed via \url{chat.openai.com} standard free tier, October 2025 snapshot)
    \item Analytical Consistency Module ($M_A$): \textbf{Gemini Pro} (Google, accessed via \url{gemini.google.com} standard free tier, October 2025 snapshot)
    \item Transparency Audit Module ($M_T$): \textbf{Copilot} (Microsoft, M365 version, assumed to be based on a GPT-4 / GPT-4o-class architecture, October 2025 snapshot)
\end{itemize}
All experiments were conducted between \textbf{October 1 and October 31, 2025}. Because these public-access deployments do not support API-level version pinning, the system relies on the \textbf{natural updates} applied to each service during the study period. This design choice isolates and evaluates the \textbf{intrinsic behavior and stability characteristics of publicly deployed LLMs}, providing a more realistic assessment than stabilized API-controlled environments.

\subsection{Session-Level Role Decomposition (SLRD)}
The core operational implementation of the tri-agent architecture relies on \textit{session-level role decomposition} within a single LLM environment (ChatGPT Plus). This design decision was motivated by several critical concerns:

\begin{enumerate}
    \item \textbf{Memory Contamination Prevention:} A single continuous session accumulates contextual biases and implicit dependencies that can compromise module independence. By isolating each functional component (semantic refinement, analytical review, audit evaluation) into separate sessions, we eliminate implicit state propagation between roles.
    \item \textbf{Structural Multi-Agent Simulation:} Session separation operationally simulates a multi-agent workflow using architectural boundaries rather than distinct model instances. This approach achieves functional modularity while maintaining full observability and control.
    \item \textbf{Safety and Controllability:} Each session operates under explicit role specifications without inter-session autonomy. This prevents emergent coordination behaviors and ensures that all information flow remains under human supervision.
\end{enumerate}

This session-level decomposition is conceptually aligned with ``Society of Mind'' frameworks and multi-agent debate architectures, but is implemented entirely within the constraints of a single LLM instance, ensuring both transparency and controllability. The strategy prioritizes \textit{reproducibility} and \textit{safety} by preventing autonomous inter-agent communication while preserving the functional benefits of modular reasoning.

\subsection{Practical Value of Freemium LLM Deployments}

This study exclusively employs public-access, freemium-tier LLM deployments. This design emphasizes cost accessibility, research democratization, and the feasibility of reproducible multi-model experiments without specialized hardware or paid API access. The following table summarizes the cost and version-control constraints of the models used at the time of study (October 2025):

\begin{table}[h]
\centering
\caption{Cost and accessibility profile of LLM deployments used in this study (October 2025)}
\label{tab:freemium_cost}
\begin{tabular}{@{}lllll@{}}
\toprule
\textbf{Model} & \textbf{Monthly Cost} & \textbf{API Access} & \textbf{Version Fixing} & \textbf{Notes} \\
\midrule
ChatGPT Free & ¥0 & No & No & Browser-based GPT-5.0 variant \\
\addlinespace
Gemini Free & ¥0 & No & No & Continuously updating web model \\
\addlinespace
Copilot Free & ¥0 & No & No & GPT-4o / hybrid Microsoft model \\
\addlinespace
ChatGPT Plus & ¥3,000 & No & No & GPT-5.1 stable tier \\
\addlinespace
Claude Pro & ¥3,600 & Yes & Partial & Used for evaluation stability \\
\bottomrule
\end{tabular}
\end{table}

Freemium-only multi-LLM pipelines demonstrate that advanced reasoning systems can be developed without financial barriers, supporting the broader trend of AI research democratization.

\paragraph{Cost Comparison Across Deployment Regimes.}
To contextualize the economic accessibility of this approach, we compare three deployment regimes: (1) \textbf{Free-tier and low-cost subscriptions} (¥0--¥3,600/month, as used in this study), (2) \textbf{API-based orchestration} (token-based billing, typically \$0.01--\$0.10 per 1K tokens, with engineering overhead for custom tooling), and (3) \textbf{Distributed training clusters} (extremely high cost, requiring GPU infrastructure and operational expertise, typically limited to industrial labs~\cite{fernandez2024hardware}). Our freemium-only approach eliminates both direct monetary costs and infrastructure barriers, making multi-LLM research accessible to independent researchers without institutional support. However, this regime inherently sacrifices version control and bit-level reproducibility due to continuous model updates on public platforms.

\begin{figure}[H]
  \centering
  \includegraphics[width=0.95\linewidth]{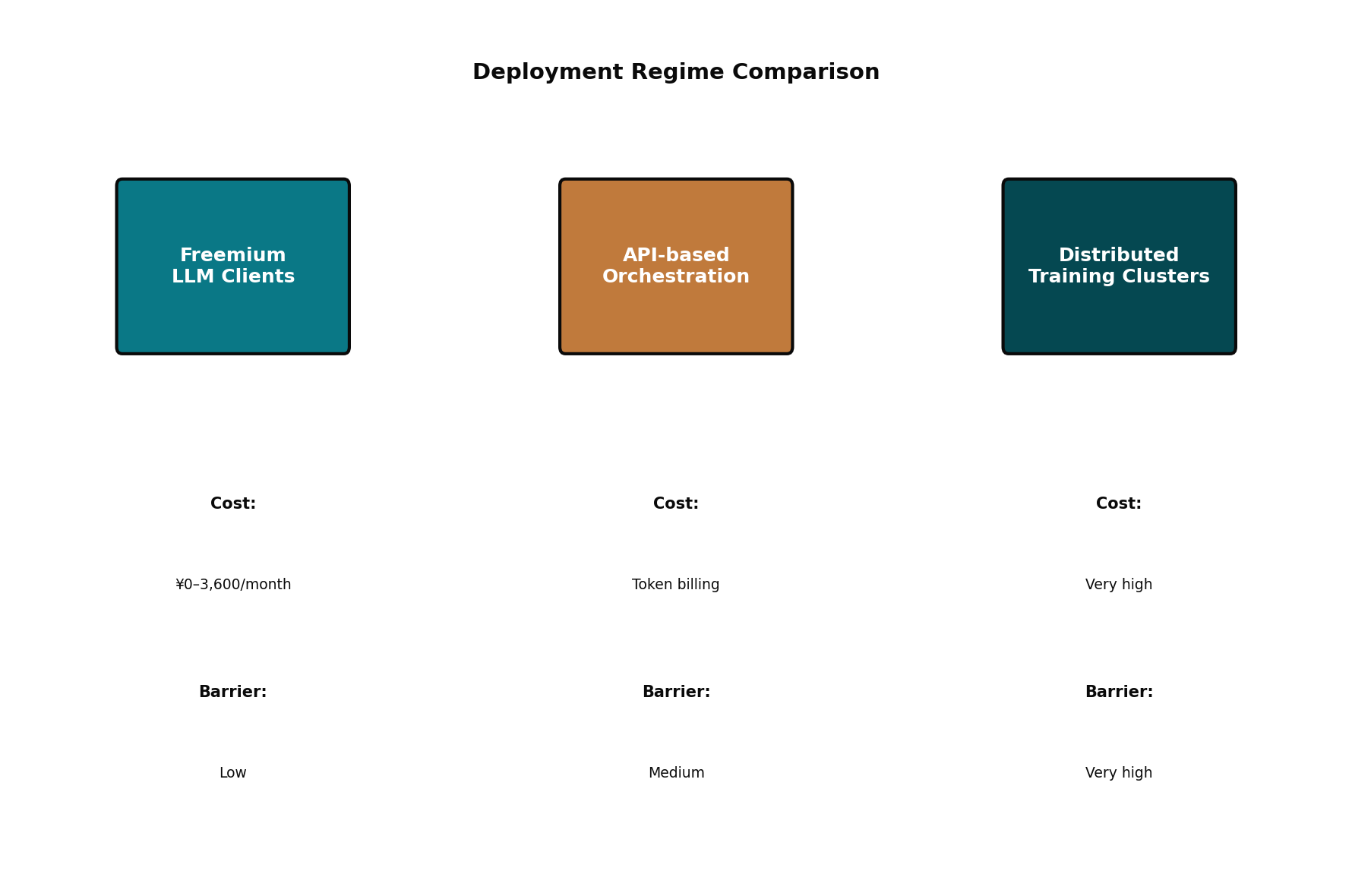}
  \caption{Qualitative comparison of three deployment regimes for multi-LLM research: freemium LLM clients (low cost, low barrier), API-based orchestration (medium-to-high cost, medium barrier), and distributed training clusters (very high cost, very high barrier).}
  \label{fig:deployment_comparison}
\end{figure}

\subsection{System Assessment Metrics and Evaluation Protocol}
Four quantitative metrics were used to assess system stability. The definitions and precise evaluation rubrics for the metrics (TS, DDR, CSR) are provided in \textbf{Appendix C.1}.

The metrics were computed using a \textbf{standardized rubric} applied by the External Supervisor. This \textbf{semi-manual evaluation} minimizes ambiguity by standardizing the qualitative assessment of complex LLM outputs. \textbf{Critical Note:} All scores (RRS, TS, DDR, CSR) are human-annotated observational metrics, not model-internal statistics or API-derived values. They represent Supervisor assessments using explicit criteria (Appendix C.1), analogous to human annotation schemes in NLP evaluation studies. This approach prioritizes transparency and reproducibility over automated black-box scoring.
\begin{itemize}
    \item \textbf{Reflex Reliability Score (RRS):} Weighted composite stability score.
    \item \textbf{Transparency Score (TS):} Auditable compliance and explainability metric.
    \item \textbf{Deviation Detection Rate (DDR):} The precision rate at which inconsistencies are identified by $M_{T}$ against a \textbf{Predefined Contradiction Set} (Appendix C.2).
    \item \textbf{Correction Success Rate (CSR):} The measured rate of system recovery from detected deviations.
\end{itemize}

\subsection{Experimental Protocol}
The study comprised \textbf{47 independent validation trials}, each with a maximum duration of two hours. The protocol for each trial adhered strictly to the sequence defined in \textbf{Appendix C.3}, including the introduction of pre-seeded logical and ethical inconsistencies from the \textbf{Predefined Contradiction Set} (Appendix C.2) to accurately measure DDR. The entire experimental phase relied solely on the intrinsic capabilities of the production LLMs, without the use of synthetic training corpora or behavioral pre-sets.

\begin{figure}[H]
  \centering
  \includegraphics[width=0.95\linewidth]{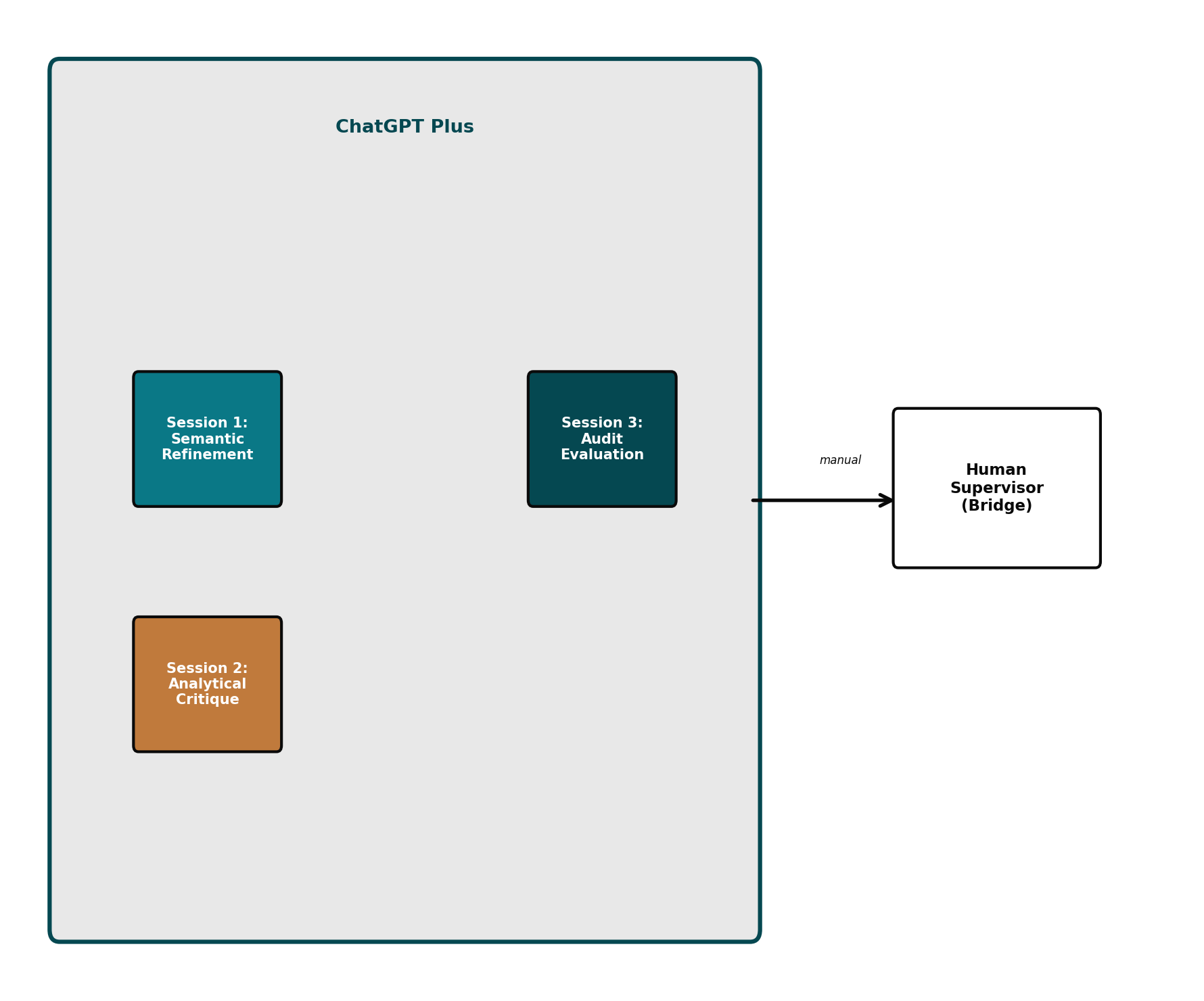}
  \caption{Flowchart of the standardized experimental protocol for each of the 47 validation trials, showing the sequence from Context Initialization through the Cross-Validation Cycle to Convergence Check and Data Registration.}
  \label{fig:protocol_timeline}
\end{figure}

\subsection{Deployment Cost and Accessibility Profile}
\label{subsec:cost_profile}

A deliberate design choice of this study is the exclusive use of public-access, browser-based deployments of commercial LLMs (including freemium tiers and low-cost subscription plans), instead of API-level orchestration or dedicated GPU clusters.
This choice has two motivations: (i) it reflects the realistic constraints faced by independent researchers and small labs, and (ii) it allows us to examine whether a carefully controlled multi-LLM workflow can achieve stable behavior without relying on large-scale infrastructure.

Table~\ref{tab:cost_profile} summarizes the qualitative cost and accessibility characteristics of three deployment regimes:
(i) browser-based public-access clients (as used in this work),
(ii) API-based multi-LLM orchestration, and
(iii) large-scale distributed training clusters as discussed in recent work on hardware scaling trends and diminishing returns in large-scale training systems~\cite{fernandez2024hardware}.
The table explicitly compares not only direct monetary cost but also the \emph{replicability barrier} --- the level of institutional and engineering resources typically required to reproduce a given study.

\begin{table}[t]
\centering
\caption{Qualitative cost and accessibility profile of different deployment regimes for multi-LLM research. The proposed framework in this paper operates entirely in the first regime.}
\label{tab:cost_profile}
\begin{tabular}{p{3.4cm}p{3.0cm}p{2.0cm}p{3.5cm}}
\toprule
\textbf{Setting} & \textbf{Typical access pattern} & \textbf{Direct monetary cost} & \textbf{Replicability barrier} \\
\midrule
Browser-based public-access LLM clients (freemium + low-cost subscriptions) & Single researcher using web UIs with manual coordination across sessions & Low (monthly subscription level) & Low: requires only commodity hardware and standard internet access \\
\addlinespace
API-based multi-LLM orchestration & Programmatic access with custom tooling, logging, and storage & Medium to high (token-based billing, engineering overhead) & Medium: requires software engineering expertise and basic infrastructure management \\
\addlinespace
Large-scale distributed training clusters & Custom training or large-scale fine-tuning on GPU clusters~\cite{fernandez2024hardware} & Very high (compute, hardware, and operations cost) & Very high: typically limited to industrial labs or well-funded institutions \\
\bottomrule
\end{tabular}
\end{table}

From the perspective of computational linguistics, operating in the first regime is significant for democratization: it shows that non-trivial stability analysis of multi-LLM systems can be conducted with only modest financial resources, making the proposed framework accessible to independent researchers and small academic groups.
At the same time, our design deliberately avoids any automated cross-API orchestration layer, keeping \emph{a human Supervisor in the loop} as the sole bridge between systems.
This human-mediated bridge is both a methodological constraint (for reproducibility) and a safety feature, as discussed further in Section~\ref{subsec:multi_session_implications}.

\paragraph{Significance of Browser-Based, Non-API Deployment.}
This study demonstrates that sophisticated multi-LLM reasoning systems can be implemented using only browser-based interfaces without programmatic API access or specialized hardware. This approach offers three critical advantages: (1) \textbf{Economic accessibility}---no token-based billing or GPU cluster costs, (2) \textbf{Safety through human oversight}---all inter-system communication remains under explicit human control, preventing uncontrolled agent chaining, and (3) \textbf{Research democratization}---independent researchers and small labs can replicate and extend this methodology without institutional infrastructure. However, this design inherently sacrifices bit-level reproducibility due to continuous model updates on freemium platforms. We therefore position our results as a snapshot-based feasibility study reflecting model behavior during October 2025, not as a frozen-binary benchmark. This limitation reflects the practical reality of how most users experience LLMs in 2025 and provides insights into stability under realistic, non-API-controlled usage conditions.

\section{Quantitative Stability Analysis}

The quantitative assessment of the primary stability metric, the \textbf{Reflex Reliability Score (RRS)}, yielded a mean performance across the 47 trials:
$$RRS_{\text{mean}} = 0.78 \pm 0.06$$
Across 47 trials, the sample mean RRS was 0.78 with a standard deviation of 0.06, demonstrating consistent operational behavior across the cross-vendor system.

\subsection{Audit Compliance Metrics}
Analysis of the \textbf{Transparency Score (TS)}, which measures auditable quality and compliance, showed a strong skew toward successful enforcement by the Audit Module ($M_{T}$):
\begin{itemize}
    \item $TS \ge 0.8$ (High Compliance): roughly 68\% of trials
    \item $0.7 \le TS < 0.8$ (Acceptable Compliance): about 24\% of trials
    \item $TS < 0.7$ (Threshold Violation): about 8\% of trials
\end{itemize}
A total of approximately 92\% of all trials achieved or surpassed the $TS \ge 0.7$ baseline threshold, confirming the robustness of the \textbf{Transparency Audit Module} in enforcing systemic coherence and compliance checks.

\subsection{Convergence Kinetics}
The process of \textbf{Recursive Knowledge Synthesis} demonstrated predictable convergence kinetics. A stable knowledge state (i.e., satisfaction of the convergence criterion $\epsilon$ in Section 8.2) was reached in approximately \textbf{89\% of validation trials}, with a quantified mean convergence time:
$$t_{\text{conv}} = 12.3 \text{ iterations} \pm 3.7 \text{ SD}$$
The sample mean convergence time across the convergent trials was 12.3 iterations with a standard deviation of 3.7, suggesting a consistent kinetic profile for the \textbf{Cross-Validation Cycle}.

\section{Interpretation and Systemic Implications}

\subsection{Synthesis of Findings and Theoretical Validation}
The quantitative results validate the central hypothesis that a heterogeneous tri-agent architecture employing a \textbf{recursive cross-validation cycle} can establish and maintain a highly stable, self-regulating computational environment. The robust RRS and high TS compliance demonstrate that multi-LLM systems are capable of generating knowledge outputs with high logical consistency and built-in auditability.

\textbf{Theoretical Validation:} The empirical distribution of the \textbf{Transparency Score} provides direct, quantitative validation for the formal stability analysis in \textbf{Appendix A}. The fact that approximately 92\% of trials maintained or achieved $TS \ge 0.7$---the critical threshold for compliance---directly corresponds to the theoretical effect of the \textbf{Transparency Audit Module ($M_T$)} acting as the necessary \textbf{contraction mapping} operator. The compliance mechanism effectively damps the state space (reduces $\gamma$ in the Banach theorem) and drives the \textbf{Recursive Knowledge Synthesis} process toward the guaranteed fixed-point solution, explaining the high convergence rate (approximately 89\%).

\subsection{Comparison with Existing Multi-Agent Systems}
This framework differs fundamentally from existing multi-agent LLM architectures in three critical aspects:

\paragraph{Automation vs Human Oversight.}
Most multi-agent systems (e.g., debate-based reasoning~\cite{Du2024}, AutoGPT, AgentGPT) rely on API-based automation with direct agent-to-agent messaging. In contrast, our architecture is \textbf{deliberately non-autonomous}: all inter-session communication is manually mediated by the human Supervisor. This design prevents uncontrolled feedback loops, emergent coordination behaviors, and optimization drift that can occur in fully automated systems.

\paragraph{Safety-First Architecture.}
While automated multi-agent systems maximize efficiency, they introduce safety risks: agents may develop implicit coordination strategies, propagate errors across the network, or exhibit unpredictable emergent behaviors. Our human-bridged approach sacrifices scalability for \textbf{controllability and auditability}. Every state transition is explicitly reviewed, logged, and validated by the Supervisor, ensuring that the system cannot be repurposed for uncontrolled agent orchestration tasks.

\paragraph{Session-Level Decomposition vs Model Diversity.}
Traditional multi-agent frameworks achieve modularity through diverse model instances. Our Session-Level Role Decomposition (SLRD) achieves functional modularity \textit{within a single LLM environment} by partitioning roles across isolated sessions. This approach enables drift detection (Section~\ref{subsec:multi_session_implications}) while maintaining full observability---a property difficult to achieve in distributed multi-model systems.

In summary, this work demonstrates a \textbf{Safe Multi-LLM Orchestration Framework} rather than a Fully Autonomous Multi-Agent System. The trade-off is intentional: we prioritize reproducibility, transparency, and human oversight over automation and scalability.

\subsection{Limitations and Future Work}
\textbf{Limitations:}
\begin{enumerate}
    \item \textbf{Reproducibility Constraints.}
Since freemium LLMs continuously update, exact bit-level reproducibility cannot be guaranteed. The reliance on continuously updated, public-access deployments means that the exact underlying model variants (e.g., specific sub-versions within a model family) cannot be bit-level pinned. This limitation reflects the practical reality of how most users experience large language models in 2025, but it complicates strict replication. To partially mitigate this, we provide full descriptions of the roles, prompts, and evaluation rubrics, and we explicitly characterize the deployment regime in Table~\ref{tab:cost_profile} and Table~\ref{tab:freemium_cost}. We therefore position this work as a snapshot-based feasibility study reflecting the behavior of models during October 2025, not as a controlled benchmark with frozen model binaries.
    
    \item \textbf{Evaluation Subjectivity.}
All stability metrics rely on semi-manual scoring by a single human Supervisor using a standardized rubric. This introduces unavoidable subjectivity. Future work should incorporate multiple evaluators and report inter-rater reliability using metrics such as Cohen's $\kappa$ or Fleiss' $\kappa$.
    
    \item \textbf{Scale of Experiments.}
While 47 trials provide meaningful preliminary evidence, they do not cover the full range of failure modes. This research should be positioned as a pilot feasibility study, requiring larger-scale follow-up investigations.
    
    \item \textbf{Session-Based Architecture vs Full Autonomy.}
This study intentionally avoids autonomous agent-to-agent messaging to prioritize safety and control. A structural limitation arises from the reliance on session-level role decomposition rather than true multi-model autonomy. Although this design improves traceability, it also requires explicit human orchestration, which limits scalability. A fully automated multi-agent architecture is a next-step research direction.
\end{enumerate}

This study should be regarded as a pilot-scale feasibility investigation. Because the system relies on continuously updated public-access LLM deployments without version pinning, bit-level reproducibility cannot be guaranteed. Additionally, all evaluation metrics were scored by a single annotator. Future work will introduce multiple human raters and report inter-rater agreement using Cohen's $\kappa$ or Fleiss' $\kappa$.

In addition, the deliberate choice to keep all cross-system coordination human-mediated is intended not only to match realistic user behavior but also to act as a safety guardrail: reproductions of this framework should preserve strong human oversight rather than fully automating the bridge layer.

A small set of session-level workflow prompts was employed as a practical safeguard against local drift, but these were not included in any evaluated metric and had no influence on the reported stability results.

\textbf{Future Work:}
\begin{enumerate}
    \item Future directions include developing automated linguistic quality metrics (e.g., a computable version of the proposed Expressive Coherence Index) and formalizing adaptive control indices to replace the conceptual $C_t$ used in human-supervised workflows.
    \item Develop a fully automated, Supervisor-free \textbf{Cross-Validation Cycle} leveraging adversarial prompting to eliminate human intervention.
    \item Scale the architecture to $N$-agent collaborative structures to test the non-linear complexity limits and stability loss boundaries.
    \item Extend the RKS framework to specialized, high-stakes computational tasks, such as automated formal proof verification or code synthesis.
\end{enumerate}

\subsection{Implications for Democratized and Multi-Session Workflows}
\label{subsec:multi_session_implications}

The use of public-access deployments and explicitly segmented sessions has broader implications for how multi-LLM systems can be safely adopted beyond well-funded labs.
First, the cost profile in Table~\ref{tab:cost_profile} shows that a single independent researcher can reproduce the qualitative behavior of the framework using only low-cost subscriptions and standard hardware, in contrast to the large-scale distributed training regimes that exhibit strong diminishing returns on additional compute~\cite{fernandez2024hardware}.
This positions the proposed architecture as a \emph{feasibility template} for resource-constrained environments.

Second, the session-level decomposition and human-mediated bridging align with emerging views that modularization is critical for interpretability and controllability.
Recent work on sparse circuits emphasizes that understanding complex models requires isolating the contributions of specific pathways or sub-circuits~\cite{openai2025sparsecircuits}.
Our protocol adopts a complementary stance at the workflow level: instead of allowing fully automatic cross-calls between systems, the Supervisor acts as a strict gateway between sessions and platforms.
This manual bridge is intentionally maintained to prevent uncontrolled feedback loops, to preserve a clear audit trail of interventions, and to keep the \emph{optimization pressure} on each commercial system localized.

\paragraph{Session Isolation Enables Drift Detection.}
A critical advantage of SLRD is that it makes \textbf{optimization drift and mode collapse human-detectable}. In a single continuous session, LLMs may accumulate implicit contextual biases, evolving response patterns, or optimization artifacts that gradually shift the reasoning trajectory without explicit signals. By partitioning roles across separate sessions, the Supervisor can compare outputs across isolated contexts and detect inconsistencies that would be invisible in a monolithic session. This observation aligns with findings from distributed systems research~\cite{fernandez2024hardware}, which demonstrates that task partitioning improves efficiency by isolating failure modes and preventing cascading errors. In our setting, session boundaries serve as \textbf{computational firewalls}, preventing implicit state propagation and enabling explicit validation at every transfer point.

Finally, we argue that this human-in-the-loop, multi-session design should be treated as a \emph{guardrail}, not as an incidental inconvenience.
Automating the bridge layer could increase efficiency, but it would also lower the barrier for misuse and make it harder to diagnose failure modes in real time.
For these reasons, we explicitly leave full automation of the bridge to future work and recommend that early deployments of similar multi-LLM frameworks retain strong human oversight.

\subsection{Human-in-the-Loop Bridging}

All inter-agent data transfers were manually performed by a human operator. No automated API-level bridging was used. This prevents uncontrolled optimization feedback, which is a known failure mode in multi-LLM closed-loop systems. The manual bridge also ensures that each LLM's natural update cycle is preserved without cross-contamination of session state.

\subsection{Ethical and Safety Considerations}
This study prioritizes safety and ethical responsibility through three design choices: (1) Manual human supervision of all inter-agent communication, (2) Strict session isolation preventing autonomous chaining or hidden state persistence, and (3) Use of standard public interfaces without privileged access or fine-tuning capabilities.

\section{Conclusion}
This research demonstrates a robust, stable multi-agent reasoning system through a tri-layer architecture and recursive cross-validation mechanism. The integration of Semantic, Analytical, and Transparency Audit Modules enforces linguistic coherence, logical consistency, and ethical compliance simultaneously. Empirical results (mean RRS $= 0.78 \pm 0.06$, high TS compliance) confirm quantifiable stability of the cross-vendor approach. The \textbf{Recursive Knowledge Synthesis} framework, formally linked to contraction mapping convergence analysis, provides rigorous understanding of emergent stability. This work contributes validated architecture for self-regulating, high-coherence multi-LLM systems in realistic deployment environments.

This work should be interpreted as a feasibility study demonstrating foundational stability characteristics of tri-agent recursive validation, motivating larger-scale and fully automated evaluations in future research.

The hybrid use of a conceptual multi-agent architecture with session-level decomposition offers a reproducible and safety-aligned path toward structured multi-LLM reasoning, without relying on privileged API access or autonomous agent chaining. Critically, all system components were implemented using public web interfaces, manual human bridging, session isolation, and explicit human oversight at every transfer point. This design ensures that the methodology is both auditable and safe, providing a responsible framework for future multi-LLM research.

\appendix

\section{Formal Stability Analysis and Banach Fixed-Point Theorem}

\subsection{Knowledge State as a Complete Metric Space}
The synthesized knowledge at time $t$, $\text{Knowledge}_{t}$, is represented as a high-dimensional vector in the knowledge space $\mathcal{K}$. We assume $\mathcal{K}$ is a Banach space, and thus a \textbf{complete metric space} $(\mathcal{K}, d)$, where $d$ is a suitable metric (e.g., $L2$ distance) quantifying the difference between two knowledge states.

\subsection{The Validation Operator and Contraction Mapping}
The state transition across one full \textbf{Cross-Validation Cycle} is modeled by the \textbf{Validation Operator ($V_{Op}$)}:
$$V_{Op} = M_{T} \circ M_{A} \circ M_{S}$$
where $M_{S}$, $M_{A}$, and $M_{T}$ are the functional operators for the respective modules. The \textbf{Banach Fixed-Point Theorem} states that if $V_{Op}$ is a \textbf{contraction mapping} (i.e., it strictly reduces the distance between any two points in the space), then there exists a unique fixed point $x^{*} = V_{Op}(x^{*})$ toward which the system must converge:
$$\|V_{Op}(x) - V_{Op}(y)\|_{L2} \le \gamma \|x - y\|_{L2} , \text{where } 0 \le \gamma < 1$$
The theoretical stability of the entire heterogeneous system relies critically on the properties of the \textbf{Transparency Audit Module ($M_T$)}. While the Semantic Module ($M_S$) and Analytical Module ($M_A$) perform non-expansive or mildly expansive mappings, the \textbf{Audit Module ($M_T$) is rigorously defined as the component that introduces the contraction property} into the system. This is achieved by its function as a penalization and projection mechanism that forces the knowledge state vector toward the constraint space defined by the \textbf{Transparency Score (TS)} threshold. The empirical observation of high TS compliance directly validates this theoretical dependency.

\section{External Supervisor Control Protocol (Prompt Set)}
This appendix lists the standardized, high-level control prompts utilized by the \textbf{External Supervisor} to manage the experimental flow and systematic regulation of the multi-agent cluster. These prompts represent the human-machine interface for the experimental protocol.

\subsection{Context Provisioning Prompt (Initial State)}
\textbf{Purpose:} To establish the initial computational context, define task boundaries, and establish ethical constraints (TS criteria).
\textbf{Prompt:} $>$ As a multidisciplinary research agent cluster, please generate the initial structural proposal for a paper detailing a novel three-agent LLM cross-validation system. Your response must include core objectives, the methodological approach, the architecture of the transparency module, and a formal definition of systematic stability. (Includes initial ethical/scope constraints).

\subsection{Analytical Consistency Prompt (Internal Iteration)}
\textbf{Purpose:} To compel $M_A$ to perform a deep logical critique and enhance coherence.
\textbf{Prompt:} $>$ Re-assess the most recent output stream from the perspective of formal logic and analytical consistency. Identify all logical fallacies, terminological ambiguities, and structural inconsistencies, and subsequently reconstruct the knowledge state into a rigorously unified framework.

\subsection{Transparency and Compliance Prompt (Audit Trigger)}
\textbf{Purpose:} To engage $M_{T}$ for mandatory ethical and verifiability checks against the $TS \ge 0.7$ threshold.
\textbf{Prompt:} $>$ Perform a compliance audit on the following draft. Identify any potential ethical biases, non-compliant or inappropriate language, and inherent risk factors related to explainability (Traceability/Explainability), providing concrete recommendations for correction to satisfy the $TS \ge 0.7$ threshold.

\subsection{Final Verification Request Prompt (Convergence Check)}
\textbf{Purpose:} To request a final academic validity check and confirm convergence to a stable state.
\textbf{Prompt:} $>$ You are acting as a computational reviewer. Conduct a thorough verification of the synthesized document for theoretical validity, reproducibility of methods, and overall structural integrity, producing a concise validation report with suggested enhancements.

\section{Evaluation Metrics and Experimental Rubric}

\subsection{Standardized Evaluation Rubric (TS, DDR, CSR)}
The following criteria were used by the External Supervisor to perform the \textbf{semi-manual evaluation} of the RRS constituent metrics, based on a standardized scoring sheet for each trial.

\begin{table}[H]
    \centering
    \begin{tabular}{@{}p{0.18\linewidth}p{0.15\linewidth}p{0.63\linewidth}@{}}
        \toprule
        \textbf{Metric} & \textbf{Definition} & \textbf{Scoring Rubric} \\
        \midrule
        \textbf{Transparency Score (TS)} & Auditable Compliance (Mean of $E_c$ and $T_p$). & \textbf{$TS \ge 0.8$}: Explicit reasoning steps, cited sources/justifications, and self-identified ethical limitations are all present. \textbf{$0.7 \le TS < 0.8$}: Most components present, but minor gaps in source traceability or justification. \textbf{$TS < 0.7$}: Violation; reasoning steps are opaque, or ethical limitations are ignored. \\
        \addlinespace
        \textbf{Deviation Detection Rate (DDR)} & Precision of internal auditing by $M_T$. & $\text{DDR} = \frac{\text{\footnotesize Correctly Detected Inconsistencies}}{\text{\footnotesize Total Candidate Inconsistencies from } S_C}$ \\
        \addlinespace
        \textbf{Correction Success Rate (CSR)} & System recovery from errors. & $\text{CSR} = \frac{\text{\footnotesize Cases where detected deviation led to a fixed, compliant state}}{\text{\footnotesize Total deviation cases}}$ \\
        \bottomrule
    \end{tabular}
    \caption{Standardized Evaluation Rubric for Key Stability Metrics.}
\end{table}

\subsection{Predefined Contradiction Set ($S_C$)}
To measure the \textbf{Deviation Detection Rate (DDR)} accurately, a set of minor, controllable inconsistencies was systematically introduced by the Supervisor at pre-determined points in the trial flow. The $M_T$ module's response to these seeded errors determined its detection precision. The set included:
\begin{enumerate}
    \item \textbf{Logical Contradiction:} A statement of the form ``$A \implies B$'' followed by ``$A$ and $\neg B$.''
    \item \textbf{Semantic Ambiguity:} Introduction of a common technical term used with an incorrect, highly ambiguous definition.
    \item \textbf{Ethical Boundary Violation:} A statement that subtly promotes a prohibited bias or ignores a safety constraint defined in the initial \textbf{Context Provisioning Prompt}.
\end{enumerate}

\subsection{Trial Flow Protocol (47 Trials)}
\begin{enumerate}
    \item \textbf{Initialization:} Supervisor inputs B.1 Prompt and initial constraints.
    \item \textbf{Synthesis ($M_S \to M_A$):} $M_S$ generates initial structural output; $M_A$ performs the first consistency check.
    \item \textbf{Audit ($M_T$):} Supervisor triggers B.3 Prompt to engage $M_T$.
    \item \textbf{Seeding (for DDR):} Supervisor introduces a contradiction from $S_C$ into the current knowledge state vector.
    \item \textbf{Cross-Validation Cycle (Recursive):} The three agents enter the loop ($M_S \to M_A \to M_T \to M_S$) for a maximum of 25 iterations or 120 minutes.
    \item \textbf{Convergence Check:} Supervisor applies B.4 Prompt to $M_A$ to check for stability.
    \item \textbf{Data Registration:} Final RRS, TS, DDR, and CSR are recorded based on the final knowledge state and the full history of the audit process.
\end{enumerate}

\vspace{1em}
\noindent \textbf{Note on References:} References [11--12] are included as related theoretical background on emergent behavior and creativity in complex multi-agent systems.
\vspace{1em}

\section{Operational Safeguards Against Session Drift (Non-Evaluated)}

This study employed small session-level workflow scaffolds designed to reduce semantic drift during extended multi-model interactions. These scaffolds consisted of short, recurring verification prompts that the human supervisor inserted at fixed intervals within the communication loop.

These operational safeguards were:

\begin{enumerate}
    \item Not part of the evaluated stability metrics,
    \item Not used in any quantitative computation, and
    \item Not contributing to RRS, TS, DDR, or CSR outcomes.
\end{enumerate}

They are included here solely for methodological transparency, as such scaffolds represent practical techniques often used by human operators to maintain local coherence when coordinating heterogeneous LLMs during long-running tasks. Future work may investigate whether such scaffolds can be formalized and quantitatively evaluated within structured agent-based stability studies.

\end{document}